\documentclass[letterpaper, 10 pt, conference]{ieeeconf}
\IEEEoverridecommandlockouts
\usepackage{multirow}
\usepackage{cite}
\usepackage{amsmath,amssymb,amsfonts}
\usepackage{algorithmic}
\usepackage{graphicx}
\usepackage{textcomp}
\usepackage{xcolor}
\def\BibTeX{{\rm B\kern-.05em{\sc i\kern-.025em b}\kern-.08em
    T\kern-.1667em\lower.7ex\hbox{E}\kern-.125emX}}
\begin{document}

\title{Are State-of-the-art Visual Place Recognition Techniques any Good for Aerial Robotics?}

\author{Mubariz Zaffar$^{1}$, Ahmad Khaliq$^{1}$, Shoaib Ehsan$^{1}$,  Michael Milford$^{2}$, Kostas Alexis$^{3}$ and Klaus McDonald-Maier$^{1}$
\thanks{This work is supported by the UK Engineering and Physical Sciences Research Council through grants EP/R02572X/1 and EP/P017487/1.}
	\thanks{$^{1}$Authors are with the Embedded and Intelligent Systems Laboratory in Computer Science and Electronic Engineering department,
		University of Essex, Colchester, United Kingdom.
		{\tt\small \{mubariz.zaffar,ahmad.khaliq,sehsan,kdm\} @essex.ac.uk}}%
	\thanks{$^{2}$Michael Milford is with the Australian Centre for Robotic Vision and School of Electrical Engineering and Computer Science, Queensland University of Technology, Brisbane, Australia.
		{\tt\small michael.milford@qut.edu.au}}
	\thanks{$^{3}$Kostas Alexis is with the Autonomous Robots Lab, University of Nevada, USA.
		{\tt\small kalexis@unr.edu}}
}

\maketitle

\begin{abstract}
Visual Place Recognition (VPR) has seen significant advances at the frontiers of matching performance and computational superiority over the past few years. However, these evaluations are performed for ground-based mobile platforms and cannot be generalized to aerial platforms. The degree of viewpoint variation experienced by aerial robots is complex, with their processing power and on-board memory limited by payload size and battery ratings. Therefore, in this paper, we collect $8$ state-of-the-art VPR techniques that have been previously evaluated for ground-based platforms and compare them on $2$ recently proposed aerial place recognition datasets with three prime focuses: a) Matching performance b) Processing power consumption c) Projected memory requirements. This gives a birds-eye view of the applicability of contemporary VPR research to aerial robotics and lays down the the nature of challenges for aerial-VPR.

\end{abstract}


\section{Introduction}
Visual Place Recognition (VPR) represents the ability of a robot to remember a previously visited place in the robot map \cite{vprasurvey}. Generally, these places are represented as a single or multiple images corresponding to nodes of the map \cite{kostavelis2015semantic}. The existing research in VPR has been focused on ground-based mobile platforms and the datasets used for evaluation contain planar viewpoint changes. However, aerial platforms like drones introduce a third dimension (vertical) to viewpoint change. This added dimension, coupled with 6-degrees of freedom of aerial platforms, limited computational payload, limited sensing payload, limited power/energy, high velocity, difficulty of local motion estimation, restrained storage and run-time memory, make VPR challenging for aerial robotics.   

While most of the datasets \cite{seqslam} \cite{merrill2018lightweight} \cite{chen2017only} \cite {sunderhauf2015place} \cite{3} \cite{4} \cite{5} \cite{nordlanddataset} \cite{8} \cite{10} used for evaluating VPR techniques have been created using cameras mounted on cars, bicycles or hand-held setups during walk;  Maffra et al. \cite{FMaffra:etal:ICRA2018} recently introduced the Shopping street datasets targeted for aerial place recognition. Therefore, in this paper, we take up the task to evaluate $8$ contemporary VPR techniques on the datasets proposed in \cite{FMaffra:etal:ICRA2018}. The objective of this paper is to answer the question: \textit {Can VPR state-of-the-art research be extended to resource-constrained aerial robotics and how can viewpoint change resulting from 6-DOF (degrees-of-freedom) platforms affect place matching performance?}  

To explain the difference between a ground-based and aerial-based platform's viewpoint variation, we show in Fig. \ref{Fig:groundvsaerialdatasets}, a comparison between existing datasets and the datasets used in our work. The novel contributions of this paper are as follows:

\begin{figure}[t]
\begin{center}
\includegraphics[width=1.0\linewidth, height=0.45\linewidth]{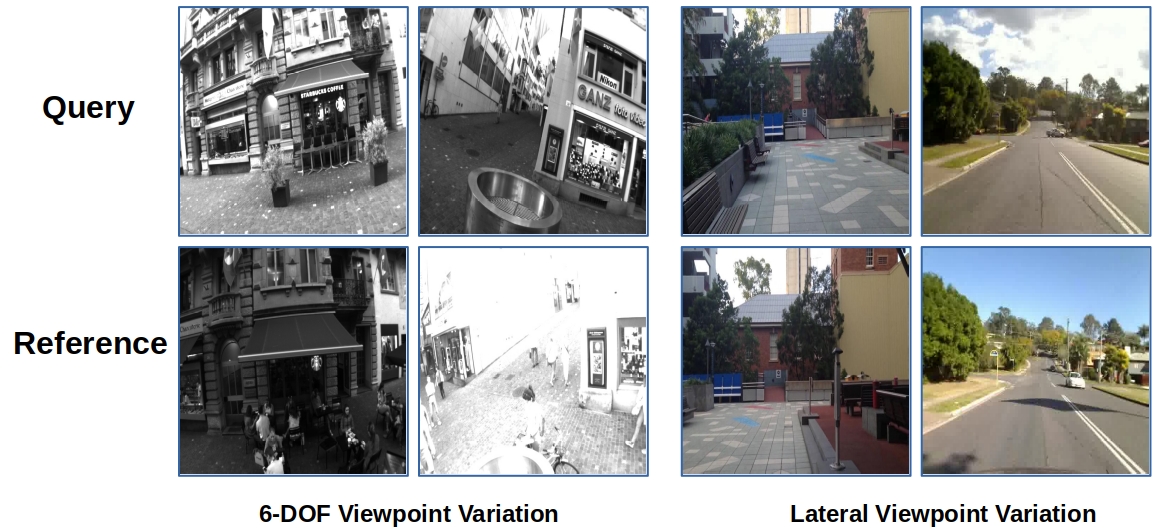}
\end{center}
\caption{The viewpoint variation challenge for aerial platforms is shown in comparison to ground based platforms. On the left, we show sample images from Shopping Street 2 dataset containing 6-DOF viewpoint change. While, on the right, we show images from the widely used Gardens Point dataset and Stlucia dataset posing lateral viewpoint change.}
\label{Fig:groundvsaerialdatasets}
\end{figure}

\begin{enumerate}
    \item {This paper discusses and evaluates inter-platform VPR (particularly ground and aerial). This is important because the performance of a VPR technique immune to planar viewpoint changes cannot be generalized to an aerial platform.}  

    \item{We present the crucial metrics of processing power needs and memory commitment to be considered at the time of selecting a VPR technique against aerial robotics. These metrics directly effect the practicality of using any VPR technique in a resource-constrained, battery powered aerial robot.}
\end{enumerate}

 It is important to note that although the datasets used in our evaluation particularly contain 3-dimensional challenging viewpoint variance, with illumination and temporal appearance change; it has been created using a hand-held rod to imitate vertical viewpoint variation. Both lateral and vertical position of the hand-held rod are continuously varied to simulate a drone-mounted camera. Despite the contrived nature of dataset, baseline VPR techniques struggle (as shown later in sub-section \ref{matching_performance_results}) from such simulated 6-DOF viewpoint variation, ``validating" the difficulty of this dataset.

The VPR techniques used for evaluation in this paper are a subset of the methods discussed in \cite{zaffar2019leveliing} and have shown promising results on different ground-based datasets. The comparison performed for aerial robotics in our paper is kept fair by deploying all the techniques on a common platform. In addition to the matching performance of all techniques, we derive the relations for processing power consumption which is an important factor of consideration for battery powered drones given limited flight-time \cite{bezzo2016online}. Unlike ground-based platforms, aerial robots are also limited by the available physical memory for storing data, one reason being the increase in payload thus, faster battery drainage. Also, larger memory size translates to greater memory power consumption. We therefore, give the projected memory requirements of all techniques for storing each dataset (feature descriptors of reference images) as a complete map. While there is significant research into compact storage of robot maps and place selection as reviewed in \cite{kostavelis2015semantic}, it is out of the scope of this work and thus, we consider all reference image descriptors as nodes of the map.  

The rest of the paper is organized as follows. Section II provides a detailed literature review of the contemporary VPR techniques used in this work. In section III, we describe the deployment configurations and experimental setup designed for analyzing the performance of VPR techniques. Section IV presents the detailed analysis and results obtained by evaluating the targeted frameworks on challenging aerial place recognition datasets. Finally, conclusions are presented is Section V.

\section{Literature Review}
Simultaneous Localization and Mapping (SLAM) represents the ability of a robot to explore and map its environment while concurrently localizing itself within it. It is one of the most challenging and well-researched domain of an autonomous robot system \cite{cadena2016past}. For autonomous systems utilizing visual sensing modalities, SLAM is discussed as visual-SLAM and the nodes of such a robot map correspond to images (or 3D scene reconstructions) of places. In a visual-SLAM system, loop-closure is achieved by matching images of the same place under different viewpoints and conditions, thus termed and surveyed as Visual Place Recognition in \cite{vprasurvey}.

Research in VPR has been the pursuit of an ultimate image-retrieval system given extreme variations since the past two decades and hence, spans the use of both handcrafted and neural networks based feature descriptors. This literature review discusses some of these techniques from the two eras of computer vision. Before the introduction of neural networks to VPR, feature extraction consisted of handcrafted feature descriptors \cite{bay2006surf} \cite{lowe2004distinctive}. Such handcrafted descriptors could further be divided into local and global feature descriptors. One of the widely known local feature descriptor is SIFT (Scale Invariant Feature Transform \cite{lowe2004distinctive}) which has been used for VPR by Stumm et al. \cite{stumm2013probabilistic}. SIFT uses Difference-of-Gaussians in scale-space to extract keypoints from an image. These keypoints are then assigned a direction of maximum change to achieve rotational-invariance and are described by oriented histogram-of-gradients. SURF (Speeded Up Robust Features) which is a modified version of SIFT was introduced by Bay et al. \cite{bay2006surf} and used in VPR by authors in \cite{murillo2007surf}. SURF has a more repeatable detector and a more distinctive descriptor than SIFT as discussed by Mistry et al. \cite{mistry2017comparison}. Other handcrafted techniques used in VPR include Centre Surrounded Extremas (CenSurE \cite{agrawal2008censure}), FAST \cite{rosten2006machine} and Bag of Visual Words (BoW \cite{sivic2003video}). 

Gist \cite{oliva2006building} is a global feature detector which uses Gabor filters to summarize the gradient information in an image and has been used for image matching by authors in \cite{murillo2009experiments} and \cite{singh2010visual}. A global variant of SURF, namely WI-SURF is used for real-time visual localization in \cite{badino2012real}. Histogram-of-oriented-gradients (HOG) \cite{MERL_TR9403} \cite{dalal2005histograms} is a standard computer vision descriptor and is used for VPR by McManus et al. in \cite{mcmanus2014scene}. Sequence of images are used in Seq-SLAM \cite{seqslam} and compared against previously visited sequences, however, it does not extract features from images but uses patch-normalized intensity frames for comparison.

Similar to the success of neural networks in different domains, VPR has seen significant advances by the use of Convolutional Neural Networks (CNNs), Convolutional Auto-Encoders (CAEs) and deep/shallow neural nets. Chen et al. discuss this in \cite{chen2014convolutional}, where given an input image to a pre-trained convolutional neural network (CNN), they extracted features from layers' responses and subsequently used these features for image comparison \cite{sermanet2013overfeat}. Following-up on their previous work, two dedicated CNNs (namely AMOSNet and HybridNet) are trained in \cite{chen2017deep} on Specific Places Dataset (SPED) achieving state-of-the-art VPR performance. Both AMOSNet and HybridNet have the same architecture as CaffeNet \cite{krizhevskyimagenet}, where the weights of former were randomly initialized while the latter used weights from CaffeNet trained on ImageNet dataset \cite{deng2009imagenet}. 

While the performance of different CNN models, training datasets and network layers has been studied for VPR; a separate paradigm of descriptor design from CNN layer (or layers) activations exists. This paradigm consists of the advent of pooling approaches employed on convolution layers including Max-Pooling \cite{tolias2015particular}, Sum-Pooling \cite{babenko2015aggregating}, Spatial Max-Pooling \cite{jaderberg2015spatial}, Cross-Pooling \cite{liu2017cross} for creating image descriptors. A CNN is intrinsically designed for classification purpose and thus the output layer consists of class labels and/or classification scores instead of image descriptors as required for VPR. Thus, Arandjelovic et al. \cite{arandjelovic2016netvlad} added a new VLAD (Vectors of Locally Aggregated Descriptors) layer to the CNN architecture which could be trained in an end-to-end manner for VPR. They subsequently plugged this VLAD layer to different CNN models and captured a highly viewpoint variant dataset from Google Street View Time Machine to train these models. However, the environmental variation seen by a neural network is limited by the unavailability of large labelled datasets of places; thus \cite{merrill2018lightweight} proposed a new unsupervised VPR-specific training mechanism. They used a convolutional auto-encoder (CAE) as the neural net machine, HOG descriptors of images were input to the CAE and the objective was to re-create the same HOG descriptor at the output layer of CAE given viewpoint and conditional variation. An interesting observation is the discussion of repetitive structures by Torii et al. \cite{torii2013visual} to propose a robust mechanism for collecting visual words into descriptors. One way to handle viewpoint variation is to acquire synthetic views of a place from different points of observation and then use these synthetic views for matching places as shown by Torii et al. \cite{torii201524}. This shows that highly conditionally variant images can still be matched given the same viewpoint and stresses on the challenges posed by viewpoint variation: the theme of this paper.

An important challenge recently has been the extraction of salient regions in an image and then using these regions of interests (ROIs) for image description to avoid confusing features. The work in \cite{tolias2015particular}, namely R-MAC (regional maximum activation
of convolutions) employs Max-Pooling over the convolutional layers' responses to encode regions. Similar to Cross-Pooling \cite{liu2017cross}, a cross-convolution technique is used to pool features from the convolutional layers by authors in \cite{chen2017only}. They first find salient region proposals from late convolutional layers of object centric VGG-16 \cite{simonyan2014very} and select top $200$ energetic regions. The regions' activations are mapped onto the previous convolutional layer, with aggregation of pre-stacked local descriptors for every mapped region. Furthermore, a regional dictionary of $10k$ words is learned from a training dataset of $5k$ images to be employed for BoW \cite{sivic2003video} thus, named as Cross-Region-BoW. Deployment on resource-limited platforms is favoured by VPR techniques that are computationally less intensive. Thus, Khaliq et al. \cite{khaliq2018holistic} proposed a lightweight CNN-based regional approach combined with VLAD (using a separate visual word vocabulary learned from a training dataset of $2.6k$ images) that has shown boost-up in image retrieval speed and accuracy. 

Although ground-based platforms have seen significant breakthroughs for visual-SLAM over the past few years, a potentially upcoming and more challenging research paradigm is visual-SLAM for aerial platforms \cite{majdik2015air}. Cieslewski et al. \cite{cieslewski2017efficient} proposed a de-centralized system that uses bag-of-words with inverted-index search for multi-robot visual place recognition. This work is followed up in \cite{cieslewski2017efficientfull}, where full-image descriptors (specifically NetVLAD) are used instead of bag-of-words for efficient de-centralized visual place recognition. This de-centralized place recognition system was combined with a de-centralized optimization system \cite{choudhary2016distributed} and a visual feature association system \cite{tardioli2015visual} by Cieslewski et al. in \cite{cieslewski2018data} to present a complete visual-SLAM system. Majdik et al. \cite{majdik2015air} present an interesting work on air-to-ground place-view projection for appearance-based urban localization of aerial vehicles. Due to the high-speed of unmanned aerial vehicles (UAVs) Vidal et al. \cite{vidal2018ultimate} combine event-cameras, having a high-dynamic range and no motion blur, with standard intensity frames and inertial measurement units to achieve an ultimate-SLAM system. 

While all of the VPR techniques available in recent literature have been evaluated for matching performance and matching time on different ground-based platforms, this paper performs a comprehensive analysis against aerial platforms. To the best of author's knowledge, this is the first work discussing and reporting the energy requirements of all of these VPR techniques to bring attention of VPR community towards energy-efficient VPR.

\section{Experimental Setup}
This section first discusses the contemporary VPR techniques that are compared in this paper. We then present the datasets used for evaluation. Finally, we describe the evaluation metrics considered for comparison in our work.    

\subsection{VPR Techniques} \label{vpr_techniques}


\subsubsection{AlexNet}
AlexNet \cite{krizhevsky2012imagenet} was introduced for image classification and achieved state-of-the-art performance on the ImageNet dataset \cite{ILSVRC15}. The applicability and performance of AlexNet for VPR was first studied by S{\"u}nderhauf et al.  \cite{krizhevsky2012imagenet}. They found \textit{conv3} to be the most robust to environmental variations. The activation maps from \textit{conv3} are encoded into feature descriptors by using Gaussian Random Projections (GRP). Our implementation of AlexNet is similar to the one presented by authors in \cite{merrill2018lightweight}.    

\subsubsection{NetVLAD}
NetVLAD \cite{arandjelovic2016netvlad} can use different CNN models because of the plug-able and train-able nature of the newly introduced VLAD layer. We have used VGG-16 \cite{simonyan2014very} as the underlying model for our NetVLAD evaluation. Pittsburgh 250K dataset is used with a dictionary size of $64$ for training the model while performing whitening on final descriptors. The utilized Python implementation has been open-sourced by \cite{cieslewski2018data}. 

\subsubsection{AMOSNet}
We implement spatial-pyramidal pooling on \textit{conv5} layer of AMOSNet to extract image descriptor from layer activations. The deployed model parameters of AMOSNet had been trained on SPED dataset and made available by authors in \cite{chen2017deep}. L1-matching is used to compare the descriptors of two images.

\subsubsection{HybridNet}
The primary difference between AMOSNet and HybridNet is that unlike AMOSNet, model weights for HybridNet are initialized from CaffeNet trained on ImageNet dataset. We use the model weights re-trained on SPED dataset and open-sourced by authors in \cite{chen2017deep}. Feature descriptor is formed by implementing spatial-pyramidal pooling on \textit{conv5} layer and L1-matching is used for matching-score computation.

\subsubsection{Cross-Region-BOW}
The MATLAB implementation for Cross-Region-BOW has been open-sourced in \cite{chen2017onlyCode}. The model used is VGG-16 that has been pre-trained on ImageNet dataset. $200$ Salient regions are identified and extracted by using \textit{conv5\_3} and \textit{conv5\_2} of the deployed model. These regions are then described using a BoW descriptor utilizing a dictionary of $10k$ words.

\subsubsection{R-MAC}
For R-MAC, we also use the VGG-16 model and extract salient regions based on maximum activations. The convolutional layer used is \textit{conv5\_2} and the implementation is inherited from \cite{tolias2015particularRMACCode}. For a fair comparison, the geometric verification block is removed. Also, power and l2 normalization is performed on the retrieved regions. Descriptors for all salient regions are cross matched at image comparison time and scores are aggregated to find the best matched image.

\subsubsection{Region-VLAD}
Convolutional layer \textit{conv4} of HybridNet is employed with $400$ ROIs in Region-VLAD. A visual word dictionary of $256$ words is used to compute the VLAD descriptor. Images are matched based on cosine-similarity of their VLAD descriptors.

\subsubsection{CALC}
For CALC (convolutional auto-encoder for loop closure), we have used the model parameters from $100,000$ iteration of the auto-encoder on Places dataset \cite{zhou2018places}. Merrill et al. have open-sourced their implementation with intrinsic AUC computation, however we only use image matches from their implementation and compute AUC as described in subsection \ref{matching_performance}.

\subsection{Evaluation Datasets}
The datasets used in our work are introduced by Maffra et al. in \cite{FMaffra:etal:ICRA2018}. Essentially the authors perform three traverses of a shopping street in the center of Zurich city from different viewpoints and create two datasets. One of the three traverses serves as a constant reference in both the datasets, while the other two traverses act as query images. Ground-truth is provided for all three traversals in the form of timestamps. The details of these datasets are summarized in Table \ref{table:datasets}.

Since the three traversals were recorded with a Visual-Inertial sensor that stores images and timestamps as ROS (Robot Operating System) bag files, we write a simple Python utility to extract images from a bag file with filenames as timestamps. We provide it here\footnote{\url{https://github.com/MubarizZaffar/rosbagextraction/}} for future ease-of-use of any datasets created using ROS-based platforms.

\renewcommand{\arraystretch}{1.2}
\renewcommand{\tabcolsep}{2.2pt}
\begin{table}[]
\caption{BENCHMARK PLACE RECOGNITION DATASETS}
\label{table:datasets}
\begin{tabular}{|c|c|c|c|c|c|}
\hline
\multirow{2}{*}{\textbf{Dataset}} & \multicolumn{2}{c|}{\textbf{Traverse}} & \multirow{2}{*}{\textbf{Environment}}                       & \multicolumn{2}{c|}{\textbf{Variation}} \\ \cline{2-3} \cline{5-6} 
                                  & \textbf{Test}    & \textbf{Reference}   &                                                             & \textbf{Viewpoint} & \textbf{Condition} \\ \hline
Shopping Street 1                          & 8577              & 7494                  & Urban                                                & moderate             & moderate        \\ \hline
Shopping Street 2                     &   4781            & 7494                  & Urban                                                       & strong        & moderate             \\ \hline
\end{tabular}
\end{table}

\subsubsection{Shopping Street 1 Dataset}
\begin{figure}[t]
\begin{center}
\includegraphics[width=1.0\linewidth, height=0.4\linewidth]{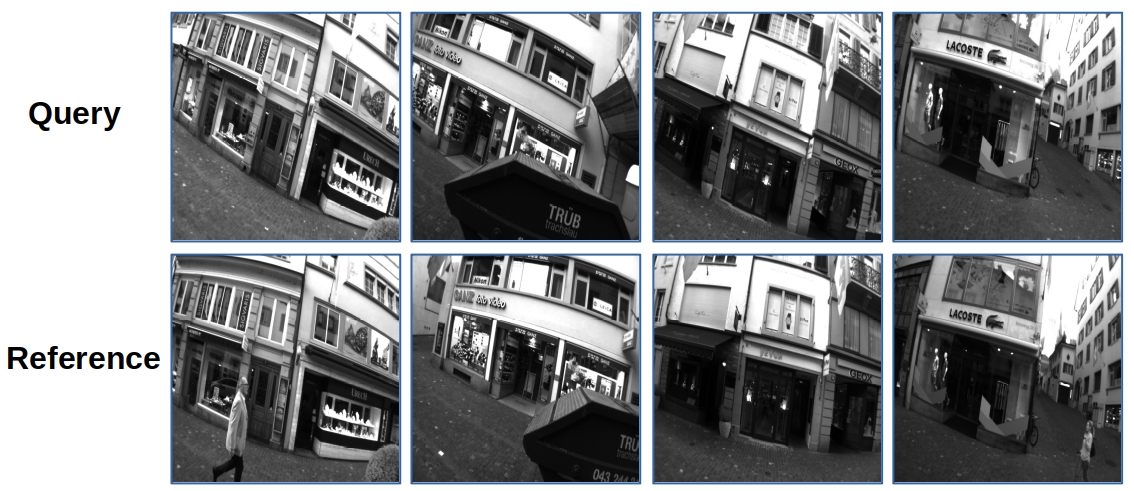}
\end{center}
\caption{Samples images from  Shopping Street 1 dataset are shown here. Top row consists of query images while the bottom row shows reference images. While this dataset contains illumination variation and dynamic objects, it does not have any extreme viewpoint variation. This makes Shopping Street 1 dataset a good reference in comparison to 6-DOF viewpoint change of Shopping Street 2 dataset (sub-section \ref{ShoppingStreet2Dataset}).}
\label{Fig:Shoppingstreetseq1vsseq2samples}
\end{figure}

This dataset consists of the two traverses of shopping street captured with a hand-held setup as shown in Fig. \ref{Fig:Shoppingstreetseq1vsseq2samples}. The undertaken traverses exhibit moderate viewpoint and appearance variation with adequate perceptual aliasing. While this dataset does not pose any significant 6-DOF viewpoint change as compared to existing VPR datasets, it serves as a good reference for the objective of this paper: observing the effect of extreme 6-DOF viewpoint change in comparison to moderate viewpoint changes. Therefore, we evaluate the $8$ state-of-the-art VPR techniques discussed in sub-section \ref{vpr_techniques} on this dataset to give a qualitative and quantitative insight into their prowess under moderate viewpoint changes.  

\subsubsection{Shopping Street 2 Dataset} \label{ShoppingStreet2Dataset}
\begin{figure}[t]
\begin{center}
\includegraphics[width=1.0\linewidth, height=0.4\linewidth]{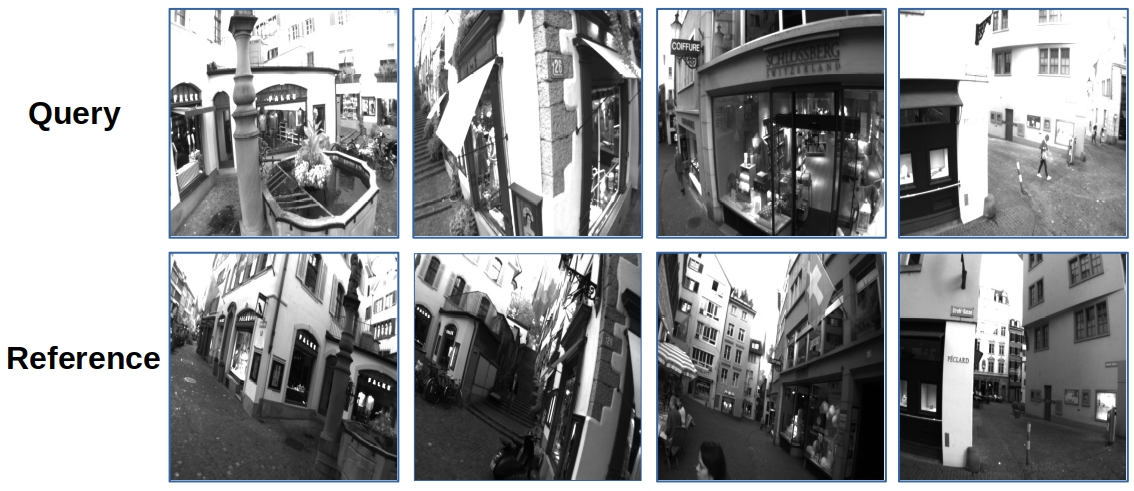}
\end{center}
\caption{Samples images from  Shopping Street 2 dataset are shown here. Top row consists of query images taken using a rod-mounted camera while bottom row shows images taken by a handheld camera. Challenging viewpoint variation is depicted here, which is very similar to the variation experienced by a 6-DOF aerial robot.}
\label{Fig:ShoppingStreet2DatasetSamples}
\end{figure}

The Shopping Street 2 dataset contains the interesting 6-DOF viewpoint change. This viewpoint change has been introduced by mounting the camera on a $4$ meter long rod such that the motion of camera imitates the flying behavior of a drone. This dataset also contains significant illumination variation and temporal appearance change. We show some sample query and reference images in Fig. \ref{Fig:ShoppingStreet2DatasetSamples}.

\subsection{Evaluation Metrics}

\subsubsection{Matching Performance} \label{matching_performance}
In image-retrieval for VPR, area under the precision-recall curves (AUC) is a well-established evaluation metric. Although, AUC has been used widely for reporting VPR performance in literature, the computational methodology used for area computation can result in different values of AUC. We compute the precision and recall values for every matched/unmatched query image. To maintain consistency in our work, we only compute and report AUC performance by utilizing equation \ref{AUC_equation}.   
\vspace{-2mm}
\begin{equation} \label{AUC_equation}
AUC = \sum_{i=1}^{N-1} \frac {(p_i + p_{i+1})} {2} \times (r_{i+1} - r_i)
\vspace{-3mm}
\end{equation}
\vspace{-2mm}

\begin{flalign*}
where; \; \; N &= No. \; of \; Query \; Images
 \\
p_i &= Precision \; at \; point \; i \\
r_i &= Recall \; at \; point \; i 
\end{flalign*}

\subsubsection{Processing Power Consumption}
The power consumption of a CPU is directly related to the CPU utilization of running processes as shown by authors in \cite{fan2007power}. Over-time, this power consumption becomes a critical factor for battery powered aerial robots. Since, computationally intense processes running for longer time-periods will quickly drain the battery, they lead to reduction of the single-charge flight-time of a drone. Therefore, we build upon the power consumption relations of \cite{fan2007power} and derive the battery expense (Ampere-hours) for each of the $8$ VPR techniques. The CPU power consumption is linked to CPU utilization by below equation \ref{power_equation}.
\vspace{-2mm}
\begin{equation} \label{power_equation}
P_c = P_i + (P_b - P_i) \times U
\vspace{-3mm}
\end{equation}

\vspace{-3mm}
\begin{flalign*}
where; \; \;
P_c &= Power \; consumption \; of \; CPU \\
P_i &= CPU \; power \; consumed \; in \; idle \; state \\
P_b &= CPU \; power \; consumed \; under \; full \; load \\
U &= CPU \; Utilization 
\end{flalign*}
\vspace{-3mm}

Given that we use the same computational platform i.e. Intel(R) Xeon(R) Gold 6134 CPU @ 3.20GHz for evaluating all $8$ VPR techniques, $P_i$ and $P_b$ can be taken as constants while $U$ is a variable parameter. Thus, by taking $P_i$ as an offset $a$ and $P_b - P_i$ as the slope $s$, equation \ref{power_equation} can further be modified as below.
\vspace{-2mm}
\begin{equation} \label{power_equation_mod}
P_c = a + (s \times U)
\end{equation}

The CPU utilization $U$ can further be broken down into the CPU utilization $U_e$ for an image feature descriptor encoding and CPU utilization $U_m$ for feature descriptor matching. These two CPU utilizations correspond to the feature encoding time $t_e$ for an input query image and query descriptor matching time $t_m$ for $M$ ($M=7494$) reference images in the database. Since encoding an input query image and matching it with all the reference images in the database is the deployment application of VPR techniques, the power consumed $P_q$ by such a process can be represented as;
\vspace{-2mm}
\begin{align}
P_q &= P_e + P_m \\
P_e &= a + (s \times U_e) \\
P_m &= a + (s \times U_m) 
\end{align}

Given that CPUs are powered from a constant voltage rail $V$ (typically $V=2.5$ volts), the ampere-hours consumed per query image $Ah_q$ can be estimated from equation \ref{energy_equation}. Thus, the total Ah consumption $Ah_t$ of each VPR technique for $N$ query images and $M$ reference images can be computed by using equation \ref{total_energy_equation}.
\vspace{-2mm}
\begin{align} 
Ah_q &= \frac  {P_e \times t_e + P_m \times t_m} {V} \label{energy_equation} \\
Ah_t &= N \times Ah_q \label{total_energy_equation}
\end{align}

\subsubsection{Projected Memory Requirement}
Although the ability to retrieve correct image matches is critical for a VPR technique, there is a trade-off between the amount of salient information that is encoded and the available on-board storage. Thus, although a VPR method can achieve excellent matching performance, its deploy-ability on an aerial platform depends on the memory footprint of its image descriptors. Therefore, for each of the $8$ VPR techniques, we provide a projected memory consumption for storing the descriptors of reference images corresponding to a complete environment traversal.

\section{Results and Analysis}
This section  discusses the performance evalutation of all the employed VPR techniques. A separate subsection is allocated to each criterion including matching performance, computational power requirements and memory usage.    

\subsection{Matching Performance} \label{matching_performance_results}
For both the benchmark datasets, this sub-section outlines and compares the AUC under PR-curves of all the $8$ VPR approaches. For a qualitative insight, we have also displayed example scenarios where query images are successfully matched or mis-matched by the employed VPR techniques. 

\subsubsection{Shopping Street 1 Dataset}
\begin{figure*}[t]
\begin{center}
\includegraphics[width=1.0\linewidth, height=0.45\linewidth]{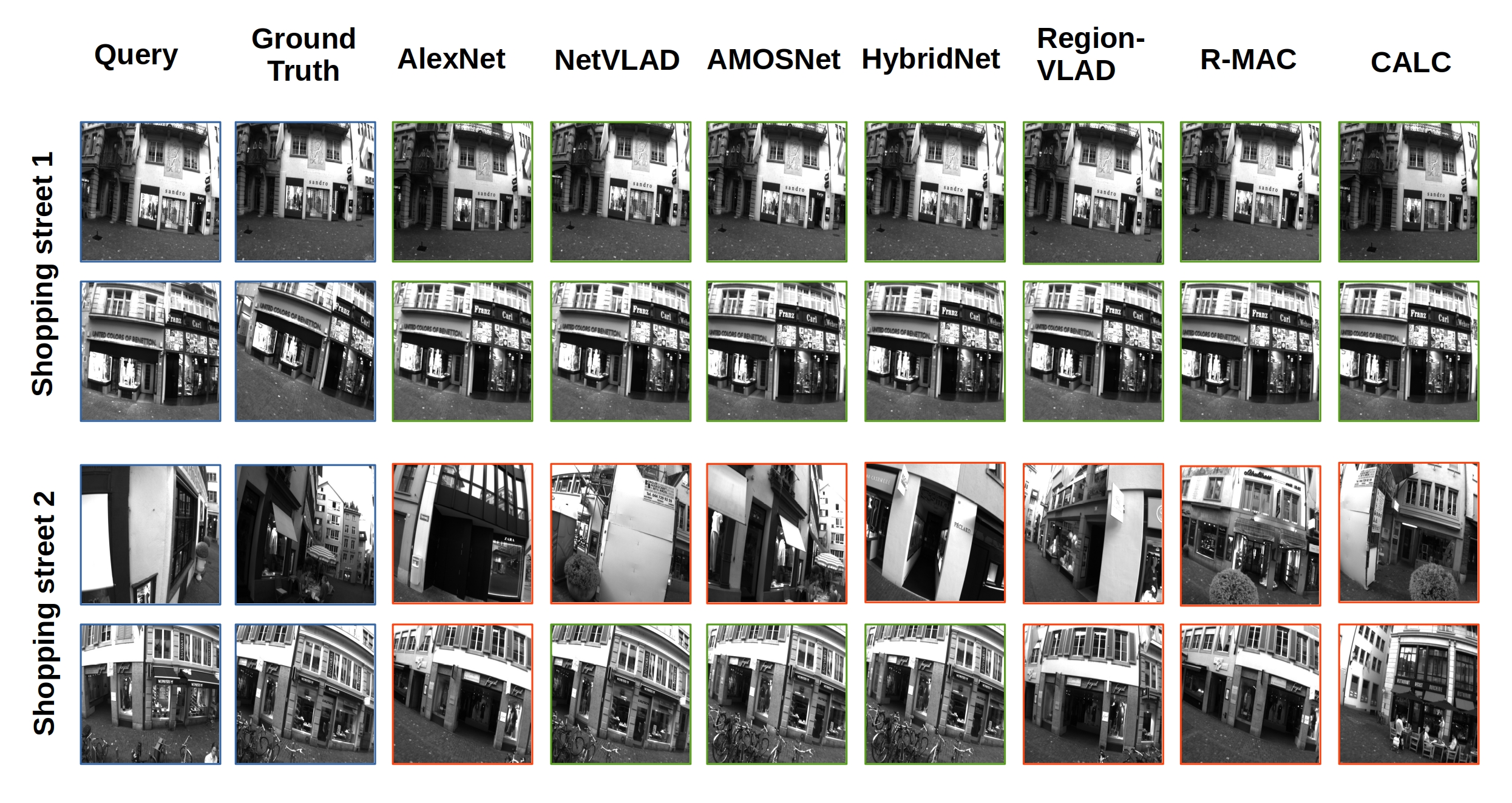}
\end{center}
\caption{Example images retrieved by all VPR techniques on both datasets are shown here. All techniques show excellent matching performance on Shopping Street 1 dataset, but struggle with 6-DOF viewpoint variation in Shopping Street 2 dataset.}
\label{Fig:ExemplarImages}
\end{figure*}

\begin{figure}[t]
\begin{center}
\includegraphics[width=1.0\linewidth, height=0.95\linewidth]{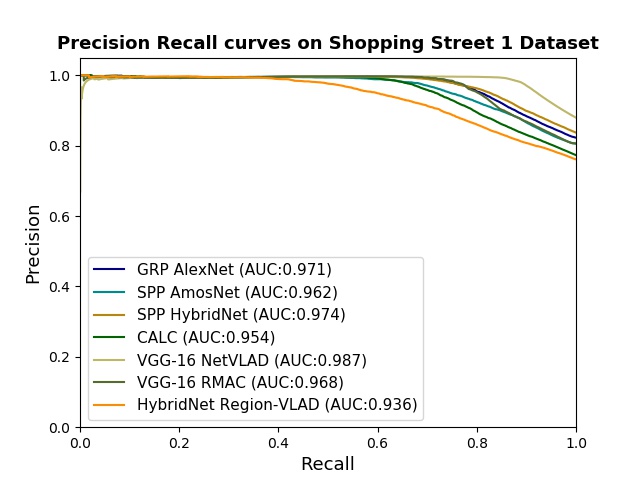}
\end{center}
\caption{AUC-PR curves of the employed VPR approaches on Shopping Street 1 Dataset are shown here. All VPR techniques achieve near-to-ideal matching performance on this dataset, advocating that the past few years of VPR research has been highly successful against planar viewpoint variations and conditional changes.}
\label{Fig:ShoppingStreet1Dataset}
\end{figure}

For this dataset, Fig. \ref{Fig:ShoppingStreet1Dataset} illustrates the PR-curves of all the employed approaches. The dataset contains mostly less-challenging planar (2-dimensional) viewpoint variation but has moderate illumination changes and occasionally observed dynamic objects, therefore, most of the techniques perform well. This shows the success of all the recently proposed VPR techniques provided moderate viewpoint variation in the dataset. Out of all the techniques and using AUC under PR-curves as an evaluation parameter, NetVLAD achieved state-of-the-art performance followed by SPP, RMAC and CALC with very minimal differences. Examples of matches/mis-matches are shown in Fig. \ref{Fig:ExemplarImages}. 

\subsubsection{Shopping Street 2 Dataset}
\begin{figure}[t]
\begin{center}
\includegraphics[width=1.0\linewidth, height=0.95\linewidth]{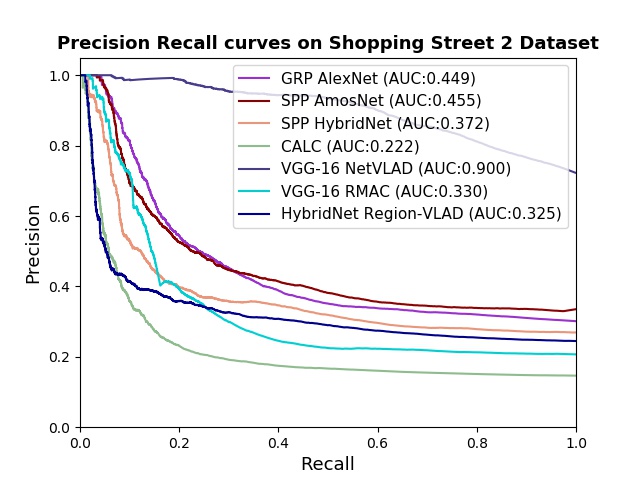}
\end{center}
\caption{AUC-PR curves of the employed VPR approaches on Shopping Street 2 Dataset are shown here. All VPR techniques clearly suffer from 6-DOF viewpoint variation in this dataset, with NetVLAD achieving state-of-the-art matching performance.}
\label{Fig:ShoppingStreet2Dataset}
\end{figure}

The query and reference traverse exhibit strong viewpoint variation in this dataset (see Fig. \ref{Fig:ShoppingStreet2DatasetSamples}). Fig. \ref{Fig:ShoppingStreet2Dataset} reports the PR curves for this dataset; showing across-the-board decline in matching performance with NetVLAD still outperforming other VPR techniques. 

The significant performance degradation for all other approaches (in comparison to NetVLAD) can be associated with the training of the CNN models. For instance, NetVLAD trained VGG-16 on an urban 250k place-centric Pittsburgh dataset exhibiting strong condition and viewpoint changes coupled with dynamic objects such as pedestrian, vehicles etc. Whereas, although HybridNet used strong condition-variant SPED dataset, the dataset intrinsically does not contain any viewpoint variation, thus, failing to perform on Shopping Street 2 dataset. Since, HybridNet is also the underlying model for Region-VLAD, where Region-VLAD does not explicitly tackle viewpoint variation; matching performance degrades under 6-DOF viewpoint change. Similarly for RMAC, VGG-16 was pre-trained on object-centric ImageNet dataset, therefore, it is not efficient in dealing with severe changes in viewpoint. Although, authors of CALC have trained their auto-encoder with viewpoint variant input images, the nature of variation is random planar projections which leads to the observed performance degradation for aerial place recognition.

In summary, a common observation across all baseline techniques is the decline of matching performance from lateral viewpoint variation of Shopping Street 1 dataset to 6-DOF viewpoint variation of Shopping Street 2 dataset. However, the trend of this decline is different between NetVLAD and the remainder techniques, primarily due to the absence of viewpoint variation in the training datasets of latter. These observations outline the need and significance of large-scale, 6-DOF viewpoint variant datasets for training VPR techniques, especially for aerial robotics.  
\vspace{-3mm}

\subsection{Processing Power Consumption}
When we talk about aerial robotics or resource-constrained platforms, energy management is the key component for any on-board deployed application. Thus, while different VPR techniques have been proposed over the years, each achieving incremental matching performance improvement and immunity to challenging conditional variations; a thorough investigation of their practicality for VPR is presented in this sub-section. We enlist in Table \ref{table:datasets}, the CPU utilization $U_e$ for feature encoding, CPU utilization $U_m$ for feature matching, feature encoding time $t_e$ and feature matching time $t_m$. By taking $a=0$, $s=1$, $V=2.5$ and $N=4781$, we also enlist the total battery consumption $Ah_t$ for all VPR techniques to give a comparative analysis. The units of times $t_e$ and $t_m$ were changed from seconds to hours for $Ah_t$ computation. Since, CPU utilization is a highly fluctuating variable therefore, we take its average over run-time of a process with a sampling rate of $0.01 sec$.    
It can be clearly seen that Cross-Region-BOW is the most power-hungry VPR technique primarily due to its computationally intense feature matching methodology. On the other hand, CALC stands-out to be the most energy-efficient technique for VPR. Please note that we do not explicitly optimize any of the VPR techniques for performance enhancement and the values of $Ah_t$ in Table \ref{table:datasets} will scale with the values of $a$ and $s$. 

\begin{table*}[]
\vspace{-4mm}
\centering
\caption{Computational Power Requirements}
\label{table:datasets}
\begin{tabular}{|c|c|c|c|c|c|c|c|c|} 
\hline
\multirow{2}{*}{\textbf{\begin{tabular}[c]{@{}l@{}}Computational\\ Performance\end{tabular}}} & \multicolumn{8}{c|}{\textbf{VPR Techniques} (Platform: Intel(R) Xeon(R) Gold 6134 CPU @ 3.20GHz with 32 cores, 64GB RAM)}                                                                                                                    \\ \cline{2-9} 
                                                                                              & \textbf{AlexNet} & \textbf{NetVLAD} & \textbf{AMOSNet} & \textbf{HybridNet} & \textbf{Cross-Region-BOW} & \textbf{R-MAC} & \textbf{Region-VLAD} & \textbf{CALC} \\ \hline
$U_e$                                                                                    & 0.734                & 0.656                & 0.437                & 0.437                  & 0.32                         & 0.5              & 0.25                    & 0.781             \\ \hline
$U_m$                                                                                 & 0.0312                & 0.036                & 0.03                & 0.03                  & 0.1                         & 0.371              & 0.031                    & 0.0312             \\ \hline
$t_e (sec)$                                                                                & 0.666                & 0.77                & 0.359                & 0.357                  & 0.834                         & 0.478              & 0.463                    & 0.027             \\ \hline
$t_m (sec)$                                                                                & 3.222                & 0.0374                & 0.614                & 0.584                  & 1199.04                         & 0.254              & 0.899                    & 0.974             \\ \hline

$Ah_t$                                                                                & 0.3128                & 0.2688                & 0.0931                & 0.0921                  & 63.836                         & 0.1768              & 0.0764                    & 0.0272             \\ \hline
\end{tabular}
\vspace{-6mm}
\end{table*}

\subsection{Projected Memory Requirement}
\begin{figure}[t]
\begin{center}
\includegraphics[width=1.0\linewidth, height=0.75\linewidth]{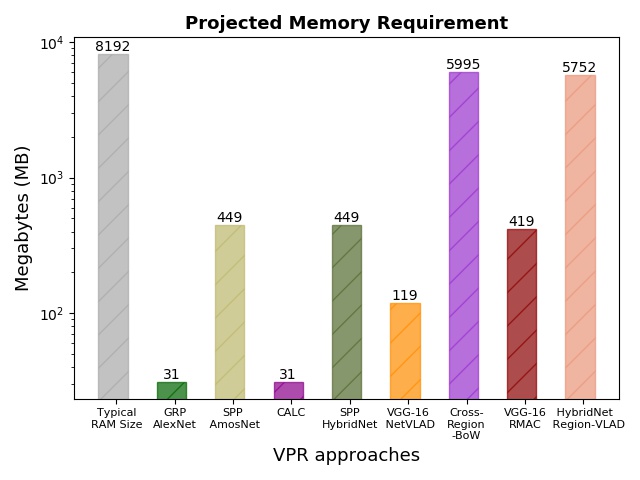}
\end{center}\vspace{-2mm}
\caption{Projected memory requirements for all the VPR approaches are shown here. The vertical axis is in logarithmic scale for clarity. The left most bar shows the typical RAM size of various development platforms. The reference features may not be loaded in RAM and could essentially be stored in an on-board SD Card which usually have similar storage capacity.}
\label{Fig:memory}
\end{figure}

One of the well-researched area in robotic navigation and mapping is the efficient storage and indexing of a robot map. This primarily involves selection of images that correspond to distinct places based on either time-interval \cite{3}, distance \cite{12}, distinctiveness \cite{13} or memorability \cite{zaffar2018memorablemaps}. While many different techniques have been presented in this regard, the underlying limitation is posed inherently by the memory footprint of a technique's image descriptor. Therefore, although place/image selection schemes can reduce the total number of images to be stored in a robot map, the size of the map will still scale with the size of a VPR technique's output descriptor. Thus, with abstraction to any place selection scheme employed, we report  in Fig. \ref{Fig:memory}, the projected memory consumption for all $8$ VPR techniques, given descriptors of all reference images are to be stored. Please note that the size of reference database for both the evaluation datasets used in this work is same and hence information in Fig. \ref{Fig:memory} is equally applicable. 
\vspace{-2mm}

\section{Conclusion}
In this paper, we performed a thorough evaluation of visual place recognition (VPR) state-of-the-art on two aerial place recognition datasets. We show that contemporary VPR techniques generally perform well on datasets containing moderate changes in viewpoint even under severe variations in illumination and conditions. However, the notable change of matching performance in between the two datasets (Shopping Street 1 and Shopping Street 2) reveals the extent of challenge posed by viewpoint variance; especially for 6-DOF (degrees-of-freedom) platforms like drones. 

We also present the limitations of VPR techniques from computational and storage perspectives given the limited on-board resources and energy supply of an aerial robot. Our evaluation is the first step into generalizability analysis of VPR techniques between different platforms and can be further extended upon proposal of more challenging aerial datasets in future. We hope this paper brings further attention towards research in energy-efficient VPR to realize real-time long-term robot autonomy. 

{
\small
\bibliographystyle{ieeetr}
\bibliography{main}
}

\end{document}